# ALIGNING A PARALLEL ENGLISH-CHINESE CORPUS STATISTICALLY WITH LEXICAL CRITERIA

Dekai Wu
*HKUST*
Department of Computer Science
University of Science & Technology
Clear Water Bay, Hong Kong
Internet: dekai@cs.ust.hk

## Abstract

We describe our experience with automatic alignment of sentences in parallel English-Chinese texts. Our report concerns three related topics: (1) progress on the HKUST English-Chinese Parallel Bilingual Corpus; (2) experiments addressing the applicability of Gale & Church's (1991) length-based statistical method to the task of alignment involving a non-Indo-European language; and (3) an improved statistical method that also incorporates domain-specific lexical cues.

## INTRODUCTION

Recently, a number of automatic techniques for aligning sentences in parallel bilingual corpora have been proposed (Kay & Röscheisen 1988; Catizone *et al.* 1989; Gale & Church 1991; Brown *et al.* 1991; Chen 1993), and coarser approaches when sentences are difficult to identify have also been advanced (Church 1993; Dagan *et al.* 1993). Such corpora contain the same material that has been translated by human experts into two languages. The goal of alignment is to identify matching sentences between the languages. Alignment is the first stage in extracting structural information and statistical parameters from bilingual corpora. The problem is made more difficult because a sentence in one language may correspond to multiple sentences in the other; worse yet, sometimes several sentences' content is distributed across multiple translated sentences.

Approaches to alignment fall into two main classes: lexical and statistical. Lexically-based techniques use extensive online bilingual lexicons to match sentences. In contrast, statistical techniques require almost no prior knowledge and are based solely on the lengths of sentences. The empirical results to date suggest that statistical methods yield performance superior to that of currently available lexical techniques.

However, as far as we know, the literature on automatic alignment has been restricted to alphabetic Indo-European languages. This methodological flaw weakens the arguments in favor of either approach, since it is unclear to what extent a technique's superiority depends on the similarity between related languages. The work reported herein moves towards addressing this problem.[1]

In this paper, we describe our experience with automatic alignment of sentences in parallel English-Chinese texts, which was performed as part of the SILC machine translation project. Our report concerns three related topics. In the first of the following sections, we describe the objectives of the HKUST English-Chinese Parallel Bilingual Corpus, and our progress. The subsequent sections report experiments addressing the applicability of a suitably modified version of Gale & Church's (1991) length-based statistical method to the task of aligning English with Chinese. In the final section, we describe an improved statistical method that also permits domain-specific lexical cues to be incorporated probabilistically.

## THE ENGLISH-CHINESE CORPUS

The dearth of work on non-Indo-European languages can partly be attributed to a lack of the prequisite bilingual corpora. As a step toward remedying this, we are in the process of constructing a suitable English-Chinese corpus. To be included, materials must contain primarily tight, literal sentence translations. This rules out most fiction and literary material.

We have been concentrating on the Hong Kong Hansard, which are the parliamentary proceedings of the Legislative Council (LegCo). Analogously to the bilingual texts of the Canadian Hansard (Gale & Church 1991), LegCo transcripts are kept in full translation in both English and Cantonese.[2] However, unlike the Canadian Hansard, the Hong Kong Hansard has not previously been available in machine-readable form. We have obtained and converted these materials by special arrangement.

---

[1] Some newer methods are also intended to be applied to non-Indo-European languages in the future (Fung & Church 1994).

The materials contain high-quality literal translation. Statements in LegCo may be made using either English or Cantonese, and are transcribed in the original language. A translation to the other language is made later to yield complete parallel texts, with annotations specifying the source language used by each speaker. Most sentences are translated 1-for-1. A small proportion are 1-for-2 or 2-for-2, and on rare occasion 1-for-3, 3-for-3, or other configurations. Samples of the English and Chinese texts can be seen in figures 3 and 4.[3]

Because of the obscure format of the original data, it has been necessary to employ a substantial amount of automatic conversion and reformatting. Sentences are identified automatically using heuristics that depend on punctuation and spacing. Segmentation errors occur occasionally, due either to typographical errors in the original data, or to inadequacies of our automatic conversion heuristics. This simply results in incorrectly placed delimiters; it does not remove any text from the corpus.

Although the emphasis is on clean text so that markup is minimal, paragraphs and sentences are marked following TEI-conformant SGML (Sperberg-McQueen & Burnard 1992). We use the term "sentence" in a generalized sense including lines in itemized lists, headings, and other non-sentential segments smaller than a paragraph.

The corpus currently contains about 60Mb of raw data, of which we have been concentrating on approximately 3.2Mb. Of this, 2.1Mb is text comprised of approximately 0.35 million English words, with the corresponding Chinese translation occupying the remaining 1.1Mb.

## STATISTICALLY-BASED ALIGNMENT

The statistical approach to alignment can be summarized as follows: choose the alignment that maximizes the probability over all possible alignments, given a pair of parallel texts. Formally, choose

$$(1) \qquad \arg\max_{\mathcal{A}} \Pr(\mathcal{A}|\mathcal{T}_1, \mathcal{T}_2)$$

where $\mathcal{A}$ is an alignment, and $\mathcal{T}_1$ and $\mathcal{T}_2$ are the English and Chinese texts, respectively. An alignment $\mathcal{A}$ is a set consisting of $L_1 \rightleftharpoons L_2$ pairs where each $L_1$ or $L_2$ is an English or Chinese passage.

This formulation is so extremely general that it is difficult to argue against its pure form. More controversial are the approximations that must be made to obtain a tractable version.

The first commonly made approximation is that the probabilities of the individual aligned pairs within an alignment are independent, i.e.,

$$\Pr(\mathcal{A}|\mathcal{T}_1, \mathcal{T}_2) \approx \prod_{(L_1 \rightleftharpoons L_2) \in \mathcal{A}} \Pr(L_1 \rightleftharpoons L_2|\mathcal{T}_1, \mathcal{T}_2)$$

The other common approximation is that each $\Pr(L_1 \rightleftharpoons L_2|\mathcal{T}_1, \mathcal{T}_2)$ depends not on the entire texts, but only on the contents of the specific passages within the alignment:

$$\Pr(\mathcal{A}|\mathcal{T}_1, \mathcal{T}_2) \approx \prod_{(L_1 \rightleftharpoons L_2) \in \mathcal{A}} \Pr(L_1 \rightleftharpoons L_2|L_1, L_2)$$

Maximization of this approximation to the alignment probabilities is easily converted into a minimum-sum problem:

$$(2)$$
$$\arg\max_{\mathcal{A}} \Pr(\mathcal{A}|\mathcal{T}_1, \mathcal{T}_2)$$
$$\approx \arg\max_{\mathcal{A}} \prod_{(L_1 \rightleftharpoons L_2) \in \mathcal{A}} \Pr(L_1 \rightleftharpoons L_2|L_1, L_2)$$
$$= \arg\min_{\mathcal{A}} \sum_{(L_1 \rightleftharpoons L_2) \in \mathcal{A}} -\log \Pr(L_1 \rightleftharpoons L_2|L_1, L_2)$$

The minimization can be implemented using a dynamic programming strategy.

Further approximations vary according to the specific method being used. Below, we first discuss a pure length-based approximation, then a method with lexical extensions.

## APPLICABILITY OF LENGTH-BASED METHODS TO CHINESE

Length-based alignment methods are based on the following approximation to equation (2):

$$(3) \quad \Pr(L_1 \rightleftharpoons L_2|L_1, L_2) \approx \Pr(L_1 \rightleftharpoons L_2|l_1, l_2)$$

where $l_1 = \text{length}(L_1)$ and $l_2 = \text{length}(L_2)$, measured in number of characters. In other words, the only feature of $L_1$ and $L_2$ that affects their alignment probability is their length. Note that there are other length-based alignment methods that measure length in number of words instead of characters (Brown *et al.* 1991). However, since Chinese text consists of an unsegmented character stream without marked word boundaries, it would not be possible to count the number of words in a sentence without first parsing it.

---

[2] Cantonese is one of the four major Han Chinese languages. Formal written Cantonese employs the same characters as Mandarin, with some additions. Though there are grammatical and usage differences between the Chinese languages, as between German and Swiss German, the written forms can be read by all.

[3] For further description see also Fung & Wu (1994).

Although it has been suggested that length-based methods are language-independent (Gale & Church 1991; Brown *et al.* 1991), they may in fact rely to some extent on length correlations arising from the historical relationships of the languages being aligned. If translated sentences share cognates, then the character lengths of those cognates are of course correlated. Grammatical similarities between related languages may also produce correlations in sentence lengths.

Moreover, the combinatorics of non-Indo-European languages can depart greatly from Indo-European languages. In Chinese, the majority of words are just one or two characters long (though collocations up to four characters are also common). At the same time, there are several thousand characters in daily use, as in conversation or newspaper text. Such lexical differences make it even less obvious whether pure sentence-length criteria are adequately discriminating for statistical alignment.

Our first goal, therefore, is to test whether purely length-based alignment results can be replicated for English and Chinese, languages from unrelated families. However, before length-based methods can be applied to Chinese, it is first necessary to generalize the notion of "number of characters" to Chinese strings, because most Chinese text (including our corpus) includes occasional English proper names and abbreviations, as well as punctuation marks. Our approach is to count each Chinese character as having length 2, and each English or punctuation character as having length 1. This corresponds to the byte count for text stored in the hybrid English-Chinese encoding system known as *Big 5*.

Gale & Church's (1991) length-based alignment method is based on the model that each English character in $L_1$ is responsible for generating some number of characters in $L_2$. This model leads to a further approximation which encapsulates the dependence to a single parameter $\delta$ that is a function of $l_1$ and $l_2$:

$$\Pr(L_1 \rightleftharpoons L_2 | L_1, L_2) \approx \Pr(L_1 \rightleftharpoons L_2 | \delta(l_1, l_2))$$

However, it is much easier to estimate the distributions for the inverted form obtained by applying Bayes' Rule:

$$\Pr(L_1 \rightleftharpoons L_2 | \delta) = \frac{\Pr(\delta | L_1 \rightleftharpoons L_2) \Pr(L_1 \rightleftharpoons L_2)}{\Pr(\delta)}$$

where $Pr(\delta)$ is a normalizing constant that can be ignored during minimization. The other two distributions are estimated as follows.

First we choose a function for $\delta(l_1, l_2)$. To do this we look at the relation between $l_1$ and $l_2$ under the generative model. Figure 1 shows a plot of English versus Chinese sentence lengths for a hand-aligned sample of 142 sentences. If the sentence lengths were perfectly correlated, the points would lie on a diagonal through the origin. We estimate the slope of this idealized diagonal $c = E(r) = E(l_2/l_1)$ by averaging over the training corpus of hand-aligned $L_1 \rightleftharpoons L_2$ pairs, weighting by the length of $L_1$. In fact this plot displays substantially greater scatter than the English-French data of Gale & Church (1991).[4] The mean number of Chinese characters generated by each English character is $c = 0.506$, with a standard deviation $\sigma = 0.166$.

We now assume that $l_2 - l_1 c$ is normally distributed, following Gale & Church (1991), and transform it into a new gaussian variable of standard form (i.e., with mean 0 and variance 1) by appropriate normalization:

$$(4) \qquad \frac{l_2 - l_1 c}{\sqrt{l_1 \sigma^2}}$$

This is the quantity that we choose to define as $\delta(l_1, l_2)$. Consequently, for any two pairs in a proposed alignment, $\Pr(\delta | L_1 \rightleftharpoons L_2)$ can be estimated according to the gaussian assumption.

To check how accurate the gaussian assumption is, we can use equation (4) to transform the same training points from figure 1 and produce a histogram. The result is shown in figure 2. Again, the distribution deviates from a gaussian distribution substantially more than Gale & Church (1991) report for French/German/English. Moreover, the distribution does not resemble any smooth distribution at all, including the logarithmic normal used by Brown *et al.* (1991), raising doubts about the potential performance of pure length-based alignment.

Continuing nevertheless, to estimate the other term $\Pr(L_1 \rightleftharpoons L_2)$, a prior over six classes is constructed, where the classes are defined by the number of passages included within $L_1$ and $L_2$. Table 1 shows the probabilities used. These probabilities are taken directly from Gale & Church (1991); slightly improved performance might be obtained by estimating these probabilities from our corpus.

The aligned results using this model were evaluated by hand for the entire contents of a ran-domly selected pair of English and Chinese files corresponding to a complete session, comprising 506 English sentences and 505 Chinese sentences. Figure 3 shows an excerpt from this output. Most of the true 1-for-1 pairs are aligned correctly. In (4), two English sentences are correctly aligned with a single Chinese sentence. However, the English sentences in (6, 7) are incorrectly aligned 1-for-1 instead of 2-for-1. Also, (11, 12) shows an example of a 3-for-1, 1-for-1 sequence that the model has no choice but to align as 2-for-2, 2-for-2.

---

[4]The difference is also partly due to the fact that Gale & Church (1991) plot paragraph lengths instead of sentence lengths. We have chosen to plot sentence lengths because that is what the algorithm is based on.

| | |
|---|---|
| 1. ¶MR FRED LI ( in Cantonese ) : | ¶李華明議員問: |
| 2. I would like to talk about public assistance. | 我想談及公共援助問題。 |
| 3. I notice from your address that under the Public Assistance Scheme, the basic rate of $825 a month for a single adult will be increased by 15% to $950 a month. | 施政報告提到提高單身人士的公共援助基本金額,由每月825元提高至950元,即加幅是15%。 |
| 4. However, do you know that the revised rate plus all other grants will give each recipient no more than $2000 a month? On average, each recipient will receive $1600 to $1700 a month. | 但你知否經過調整後,即使加上所有其他津貼,每名受助者每月所得到的公共援助都不會超過2000元,平均來說,他們每月所得的是1600元至1700元左右。 |
| 5. In view of Hong Kong's prosperity and high living cost, this figure is very ironical. | 以香港的繁榮和生活水平之高,這數字根本是一個很大的諷刺。 |
| 6. May I have your views and that of the Government? | 請問政府或總督先生,你有何看法,是否覺得應全面檢討公共援助的計算方式? |
| 7. Do you think that a comprehensive review should be conducted on the method of calculating public assistance? | 因為基數這麼低,就算加多20%至30%,仍是遠遠落後於現時的生活水平。 |
| 8. Since the basic rate is so low, it will still be far below the current level of living even if it is further increased by 20% to 30%. If no comprehensive review is carried out in this aspect, this " safety net " cannot provide any assistance at all for those who are really in need. | 若不全面檢討公共援助的計算方法,這安全網根本不能為那些真正有需要的人士提供協助。 |
| 9. I hope Mr Governor will give this question a serious response. | 希望總督先生認真回應這問題。 |
| 10. ¶THE GOVERNOR: | ¶總督答(譯文): |
| 11. It is not in any way to belittle the importance of the point that the Honourable Member has made to say that, when at the outset of our discussions I said that I did not think that the Government would be regarded for long as having been extravagant yesterday, I did not realize that the criticisms would begin quite as rapidly as they have. | 我在昨天的討論開始時說,我相信政府不會長期被指為揮霍無度。 當時我沒有料到批評會來得這麼快。 |
| 12. The proposals that we make on public assistance, both the increase in scale rates, and the relaxation of the absence rule, are substantial steps forward in Hong Kong which will, I think, be very widely welcomed. | 我說這句話,絕對無意貶低這位議員剛才所提意見的重要性。 我們就公共援助提出的建議,不論是增加援助金額或是放寬離港期限的規定,對本港來說,可說是向前跨進一大步,我相信普遍會受到歡迎。 |
| 13. But I know that there will always be those who, I am sure for very good reason, will say you should have gone further, you should have done more. | 不過,我知道有些人一定會說,你應更向前邁進一步,你應該做多一些,我肯定他們這樣說是有非常充分的理由。 |
| 14. Societies customarily make advances in social welfare because there are members of the community who develop that sort of case very often with eloquence and verve. | 很多社會慣於改善其社會福利,原因是有些人經常利用動聽的說話及凌厲的詞鋒,提出這方面的意見。 |

Figure 3: A sample of length-based alignment output.

Judging relative to a manual alignment of the English and Chinese files, a total of 86.4% of the true $L_1 \rightleftharpoons L_2$ pairs were correctly identified by the length-based method. However, many of the errors occurred within the introductory session header, whose format is domain-specific (dis-cussed below). If the introduction is discarded, then the proportion of correctly aligned pairs rises to 95.2%, a respectable rate especially in view of the drastic inaccuracies in the distributions assumed. A detailed breakdown of the results is shown in Table 2. For reference, results reported for English/French generally fall between 96% and 98%. However, all of these numbers should be interpreted as highly domain dependent, with very small sample size.

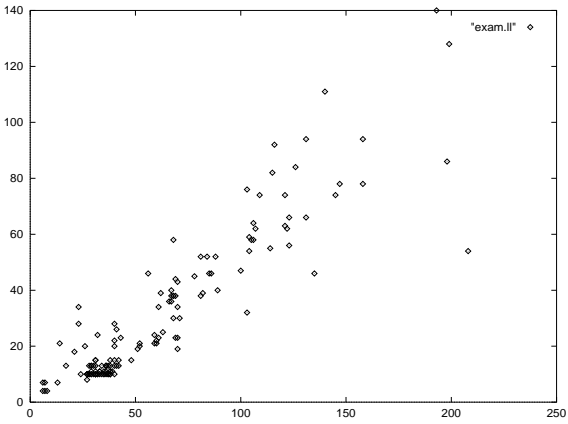

Figure 1: English versus Chinese sentence lengths.

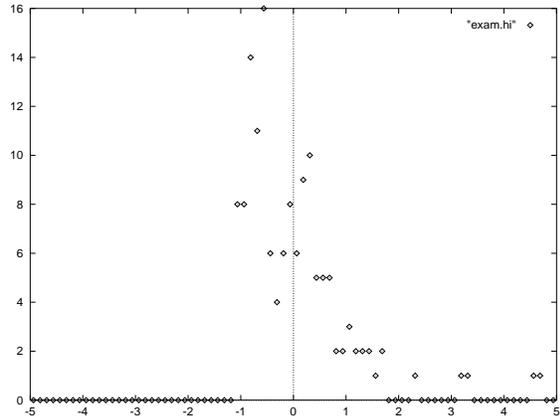

Figure 2: English versus Chinese sentence lengths.

| # segments | | $\Pr(L_1 \rightleftharpoons L_2)$ |
|---|---|---|
| $L_1$ | $L_2$ | |
| 0 | 1 | 0.0099 |
| 1 | 0 | 0.0099 |
| 1 | 1 | 0.89 |
| 1 | 2 | 0.089 |
| 2 | 1 | 0.089 |
| 2 | 2 | 0.011 |

Table 1: Priors for $\Pr(L_1 \rightleftharpoons L_2)$.

The above rates are for Type I errors. The alternative measure of accuracy on Type II errors is useful for machine translation applications, where the objective is to extract only 1-for-1 sentence pairs, and to discard all others. In this case, we are interested in the proportion of 1-for-1 *output* pairs that are true 1-for-1 pairs. (In information retrieval terminology, this measures precision whereas the above measures recall.) In the test session, 438 1-for-1 pairs were output, of which 377, or 86.1%, were true matches. Again, however, by discarding the introduction, the accuracy rises to a surprising 96.3%.

The introductory session header exemplifies a weakness of the pure length-based strategy, namely, its susceptibility to long stretches of passages with roughly similar lengths. In our data this arises from the list of council members present and absent at each session (figure 4), but similar stretches can arise in many other domains. In such a situation, two slight perturbations may cause the entire stretch of passages between the perturbations to be misaligned. These perturbations can easily arise from a number of causes, including slight omissions or mismatches in the original parallel texts, a 1-for-2 translation pair preceding or following the stretch of passages, or errors in the heuristic segmentation preprocessing. Substantial penalties may occur at the beginning and ending boundaries of the misaligned region, where the perturbations lie, but the misalignment between those boundaries incurs little penalty, because the mismatched passages have apparently matching lengths. This problem is apparently exacerbated by the non-alphabetic nature of Chinese. Because Chinese text contains fewer characters, character length is a less discriminating feature, varying over a range of fewer possible discrete values than the corresponding English. The next section discusses a solution to this problem.

In summary, we have found that the statistical correlation of sentence lengths has a far greater variance for our English-Chinese materials than with the Indo-European materials used by Gale & Church (1991). Despite this, the pure length-based method performs surprisingly well, except for its weakness in handling long stretches of sentences with close lengths.

## STATISTICAL INCORPORATION OF LEXICAL CUES

To obtain further improvement in alignment accuracy requires matching the passages' lexical content, rather than using pure length criteria. This is particularly relevant for the type of long mismatched stretches described above.

Previous work on alignment has employed ei-

|        | 1-1  | 1-2  | 2-1  | 2-2 | 1-3 | 3-1 | 3-3 |
|--------|------|------|------|-----|-----|-----|-----|
| Total  | 433  | 20   | 21   | 2   | 1   | 1   | 1   |
| Correct| 361  | 17   | 20   | 0   | 0   | 0   | 0   |
| Incorrect | 11 | 3   | 1    | 2   | 1   | 1   | 1   |
| % Correct | 87.1 | 85.0 | 95.2 | 0.0 | 0.0 | 0.0 | 0.0 |

Table 2: Detailed breakdown of length-based alignment results.

1. ¶THE DEPUTY PRESIDENT THE HONOURABLE JOHN JOSEPH SWAINE, C.B.E., Q.C., J.P. ⌋     ¶布政司霍德爵士議員, K.B.E., L.V.O., J.P. ⌋
2. ¶THE CHIEF SECRETARY THE HONOURABLE SIR DAVID ROBERT FORD, K.B.E., L.V.O., J.P. ⌋     ¶財政司麥高樂議員, C.B.E., J.P. ⌋
3. ¶THE FINANCIAL SECRETARY THE HONOURABLE NATHANIEL WILLIAM HAMISH MACLEOD, C.B.E., J.P. ⌋     ¶律政司馬富善議員, C.M.G., J.P. ⌋

⋮ *37 misaligned matchings omitted*

41. ¶THE HONOURALBE MAN SAI - CHEONG ⌋     ¶潘國濂議員 ⌋
42. ¶THE HONOURABLE STEVEN POON KWOK - LIM THE HONOURABLE HENRY TANG YING - YEN, J.P. ⌋     ¶唐英年議員, J.P. ⌋
43. ¶THE HONOURABLE TIK CHI - YUEN ⌋     ¶狄志遠議員 ⌋

Figure 4: A sample of misalignment using pure length criteria.

ther solely lexical or solely statistical length criteria. In contrast, we wish to incorporate lexical criteria without giving up the statistical approach, which provides a high baseline performance.

Our method replaces equation (3) with the following approximation:

$$\Pr(L_1 \rightleftharpoons L_2 | L_1, L_2)$$
$$\approx \Pr(L_1 \rightleftharpoons L_2 | l_1, l_2, v_1, w_1, \ldots, v_n, w_n)$$

where $v_i = \#\text{occurrences}(\text{English cue}_i, L_1)$ and $w_i = \#\text{occurrences}(\text{Chinese cue}_i, L_2)$. Again, the dependence is encapsulated within difference parameters $\delta_i$ as follows:

$$\Pr(L_1 \rightleftharpoons L_2 | L_1, L_2)$$
$$\approx \Pr(\ L_1 \rightleftharpoons L_2 |$$
$$\delta_0(l_1, l_2), \delta_1(v_1, w_1), \ldots, \delta_n(v_n, w_n))$$

Bayes' Rule now yields

$$\Pr(L_1 \rightleftharpoons L_2 | \delta_0, \delta_1, \delta_2, \ldots, \delta_n)$$
$$\propto \Pr(\delta_0, \delta_1, \ldots, \delta_n | L_1 \rightleftharpoons L_2) \Pr(L_1 \rightleftharpoons L_2)$$

The prior $\Pr(L_1 \rightleftharpoons L_2)$ is evaluated as before. We assume all $\delta_i$ values are approximately independent, giving

(5)
$$\Pr(\delta_0, \delta_1, \ldots, \delta_n | L_1 \rightleftharpoons L_2) \approx \prod_{i=0}^{n} \Pr(\delta_i | L_1 \rightleftharpoons L_2)$$

The same dynamic programming optimization can then be used. However, the computation and memory costs grow linearly with the number of lexical cues. This may not seem expensive until one considers that the pure length-based method only uses resources equivalent to that of a single lexical cue. It is in fact important to choose as few lexical cues as possible to achieve the desired accuracy.

Given the need to minimize the number of lexical cues chosen, two factors become important. First, a lexical cue should be highly reliable, so that violations, which waste the additional computation, happen only rarely. Second, the chosen lexical cues should occur frequently, since computing the optimization over many zero counts is not useful. In general, these factors are quite domain-specific, so lexical cues must be chosen for the particular corpus at hand. Note further that when these conditions are met, the exact probability distribution for the lexical $\delta_i$ parameters does not have much influence on the preferred alignment.

The bilingual correspondence lexicons we have employed are shown in figure 5. These lexical items are quite common in the LegCo domain. Items like "C.B.E." stand for honorific titles such as "Commander of the British Empire"; the other cues are self-explanatory. The cues nearly always appear 1-to-1 and the differences $\delta_i$ therefore have a mean of zero. Given the relative unimportance of the exact distributions, all were simply assumed to be normally distributed with a variance of 0.07 instead of sampling each parameter individually. This variance is fairly sharp, but nonetheless, conservatively reflects a lower reliability than most of the cues actually possess.

|          |          |          |          |          |          |
|----------|----------|----------|----------|----------|----------|
|          |          | :        | :        |          |          |
|          |          | governor | 總督      |          |          |

| C.B.E. | C.B.E. | C.M.G. | C.M.G. | I.S.O. | I.S.O. |
|--------|--------|--------|--------|--------|--------|
| J.B.E. | J.B.E. | J.P.   | J.P.   | K.B.E. | K.B.E. |
| L.V.O. | L.V.O. | O.B.E. | O.B.E. | M.B.E. | M.B.E. |
| Q.C.   | Q.C.   | January | 一月 | February | 二月 |
| March | 三月 | April | 四月 | May | 五月 |
| June | 六月 | July | 七月 | August | 八月 |
| September | 九月 | October | 十月 | November | 十一月 |
| December | 十二月 | Monday | 星期一 | Tuesday | 星期二 |
| Wednesday | 星期三 | Thursday | 星期四 | Friday | 星期五 |
| Saturday | 星期六 | Sunday | 星期日 |  |  |

Figure 5: Lexicons employed for paragraph (top) and sentence (bottom) alignment.

Using the lexical cue extensions, the Type I results on the same test file rise to 92.1% of true $L_1 \rightleftharpoons L_2$ pairs correctly identified, as compared to 86.4% for the pure length-based method. The improvement is entirely in the introductory session header. Without the header, the rate is 95.0% as compared to 95.2% earlier (the discrepancy is insignificant and is due to somewhat arbitrary decisions made on anomalous regions). Again, caution should be exercised in interpreting these percentages.

By the alternative Type II measure, 96.1% of the output 1-for-1 pairs were true matches, compared to 86.1% using the pure length-based method. Again, there is an insignificant drop when the header is discarded, in this case from 96.3% down to 95.8%.

## CONCLUSION

Of our raw corpus data, we have currently aligned approximately 3.5Mb of combined English and Chinese texts. This has yielded 10,423 pairs classified as 1-for-1, which we are using to extract more refined information. This data represents over 0.217 million English words (about 1.269Mb) plus the corresponding Chinese text (0.659Mb).

To our knowledge, this is the first large-scale empirical demonstration that a pure length-based method can yield high accuracy sentence alignments between parallel texts in Indo-European and entirely dissimilar non-alphabetic, non-Indo-European languages. We are encouraged by the results and plan to expand our program in this direction.

We have also obtained highly promising improvements by hybridizing lexical and length-based alignment methods within a common statistical framework. Though they are particularly useful for non-alphabetic languages where character length is not as discriminating a feature, we believe improvements will result even when applied to alphabetic languages.

## ACKNOWLEDGEMENTS

I am indebted to Bill Gale for helpful clarifying discussions, Xuanyin Xia and Wing Hong Chan for assistance with conversion of corpus materials, as well as Graeme Hirst and Linda Peto.